\documentclass[sigconf]{acmart}
\usepackage{subcaption}  
\usepackage{listings}
\usepackage[utf8]{inputenc}
\usepackage{array} 
\usepackage{adjustbox}
\usepackage{tabularx}
\usepackage{placeins}
\usepackage[table]{xcolor}

\AtBeginDocument{%
  }

\setcopyright{acmlicensed}
\copyrightyear{2025}
\acmYear{2025}
\acmDOI{}
\acmConference[KDDCup' 25]{In Proceedings of 2025 KDD Cup Workship for Multimodal Retrieval Augmented Generation at the 31th ACM SGKDD Conference on Knowledge Discovery and Data Mining (KDDCup' 25)}{August 03--07,
  2025}{Toronto, ON, Canada}
\acmISBN{}

\begin{document}

\title{GroundSight: Augmenting Vision-Language Models with Grounding Information and De-hallucination}


\author{Xinxi Chen}
\authornote{This work builds on a prior project at Stanford University, with sincere thanks for the support.}
\affiliation{Independent Researcher\country{}}
\email{xc336@cornell.edu}

\author{Tianyang Chen}
\authornotemark[1]

\affiliation{Independent Researcher\country{}}
\email{tiche@seas.upenn.edu}

\author{Lijia Hong}
\affiliation{Independent Researcher\country{}}
\email{lhong@alumni.ust.hk}

\renewcommand{\shortauthors}{Chen et al.}

\begin{abstract}
We propose a method to improve Visual Question Answering (VQA) with Retrieval-Augmented Generation (RAG) by introducing text-grounded object localization. Rather than retrieving information based on the entire image, our approach enables the model to generate a bounding box around the object most relevant to the question, allowing for targeted image cropping and focused retrieval. This reduces background noise, improves alignment between visual and textual cues, and helps mitigate hallucinations. Our RAG method enhances context-aware VQA responses increased the accuracy from 22.19\% to 25.64\%, with an absolute increase of 3.45 percentage points, compared to the baseline Llama-3.2-Vision-11B agent. We also proposed a de-hallucination method based on question type which can effectively reduce the hallucination rate from 65.79\% to 13.88\% and improves the truthfulness score.
\end{abstract}




\maketitle

\section{Introduction}

Visual Question Answering (VQA) \cite{Agrawal2015VQA} sits at the intersection of computer vision and natural language processing, requiring systems to reason over both images and text to produce meaningful answers. Recent advances in Vision-Language Models (VLMs) \cite{bordes2024introduction} have significantly enhanced the ability of machines to jointly understand visual and linguistic content, enabling more accurate and context-aware interpretations of complex visual scenes. However, these models are inherently limited by the knowledge encoded in their training data. To address this, Retrieval-Augmented Generation (RAG) \cite{lewis2020retrieval} introduces an external knowledge retrieval step that grounds model outputs in up-to-date or domain-specific information, bridging the gap between perception and world knowledge. The combination of VLMs with RAG is particularly important for VQA, as it allows systems not only to interpret what they see, but also to reason with additional contextual or factual information—ultimately leading to more robust, informed, and trustworthy responses. In this paper we investigate methods that can improve VQA performance.

A core challenge in combining VQA with RAG lies in identifying and retrieving external knowledge that is simultaneously relevant to both the textual query and the visual content. Unlike pure text-based RAG, where the query alone guides document selection, multimodal RAG must interpret features from an image—objects, scenes, spatial relationships—and align them with the user’s question to construct a precise retrieval request. For example, when give the question "How much does this cost?" and an image of users holding one object in hand, while the background contains multiple other objects, the VLM must be able to understand which object is the question referring to in order to perform effective informational retrieval. If the retrieval system focuses too narrowly on one modality (e.g., only on keywords in the text), it risks ignoring critical visual cues; conversely, overemphasizing visual attributes may surface facts that have little bearing on the question’s intent. Moreover, the retrieved facts must be filtered and integrated in a way that respects the visual context—misaligned or tangential information can lead to confident but incorrect answers. Balancing these two streams of information to surface grounded, image-aware knowledge is therefore a delicate orchestration that remains an open research frontier in multimodal language understanding.

Furthermore, dehallucinating VLMs is critical to ensuring the reliability and safety of their outputs, especially in applications like VQA where users may rely on responses for decision-making. VLMs, like their language-only counterparts, are prone to hallucination—generating plausible but factually incorrect or visually inconsistent answers—when they lack sufficient understanding or context. This issue is exacerbated when VLMs must reason about complex scenes or incorporate external knowledge, as they may confidently assert false claims not supported by the image or retrieved content. For the example provided by the previous section, a better answer is "I don't know" rather than responding with wrong price for the wrong object.

\section{Related Work}
Improving VQA has been a dynamic area of research, with recent efforts focusing on enhancing the gorunding of visual and textual information. GLIP \cite{Li_2022_CVPR} (Grounded Language-Image Pre-training) is a unified vision-language architecture designed to bridge object detection and phrase grounding by reformulating object detection as a vision-language matching task. The core idea is to align image regions (bounding boxes) with phrases from a natural language prompt, enabling the model to detect and ground objects based on open-vocabulary textual queries rather than a fixed set of class labels. GLIP handles both object detection and phrase grounding with a single architecture allowing the model to localize objects of interest that can best answer a given question prompt. Similarly, Grounding DINO \cite{liu2023groundingdino} tightly integrates language and vision to enable detection and localization of arbitrary objects specified by natural language prompts, rather than being limited to a fixed set of classes. Its architecture features dual backbones for image and text, a feature enhancer module that deeply fuses visual and linguistic features via cross-attention, a language-guided query selection mechanism that dynamically chooses relevant image regions based on the prompt, and a cross-modality decoder that refines predictions by alternating attention between image and text features. Grounding DINO is designed to handle referring expressions in text prompts—including pronouns like "it"—as part of its referring expression comprehension (REC) capability. The model can localize and identify specific objects or regions within an image based on a given textual description, which may include coreferences such as "it" if the context in the prompt is clear enough for the model to resolve what "it" refers to, which is crucial for targeted retrieval augmentation for VQA. The "Chain-of-Spot" \cite{dong2024chainofspot} introduces a novel and efficient approach to enhancing the visual reasoning capabilities of large vision-language models (LVLMs) through an interactive reasoning process. What sets Chain-of-Spot apart is its focus on dynamically identifying and attending to key regions of interest (ROI) within an image that are most relevant to the posed question or instruction, rather than processing the entire image at a fixed (often low) resolution. This is achieved by prompting the model to first localize the critical region in response to a query, cropping or zooming in on that region, and then generating the answer based on both the original and the focused image. This interactive, multi-step procedure allows the model to access more detailed and multi-granularity visual features without increasing the computational cost associated with higher-resolution processing.

Visual grounding aims to localize the image region referred to by a given language expression. Methodologies in this field have evolved from multi-modal fusion over a fixed set of detected objects to direct bounding-box prediction with open-vocabulary capabilities. Early approaches integrated object-level visual features into textual representations to enhance generic VQA performance (e.g., via object-text fusion strategies \cite{alberti-etal-2019-fusion}). Later work introduced more structured two-stage pipelines: for instance, a "Locate Then Generate" framework first predicts the relevant scene-text region and then generates the answer from the cropped area \cite{zhu2023locategeneratebridgingvision}. Recent efforts extend this paradigm to the video domain, where grounding scene-text temporally across frames proves beneficial for text-based video QA \cite{zhou2025scenetextgroundingtextbasedvideo}.

\newpage

\section{Proposed Method: GroundSight}
We aim to improve Visual Question Answering (VQA) in real-world scenarios, where ambiguous questions and cluttered backgrounds make it difficult for vision-language models to focus on the correct visual content. To address this, we propose GroundSight, a retrieval-augmented system that localizes the object of interest and mitigates hallucination.

\begin{figure*}[htbp]
  \centering
  \begin{subfigure}{0.48\textwidth}
    \centering
    \includegraphics[width=\linewidth]{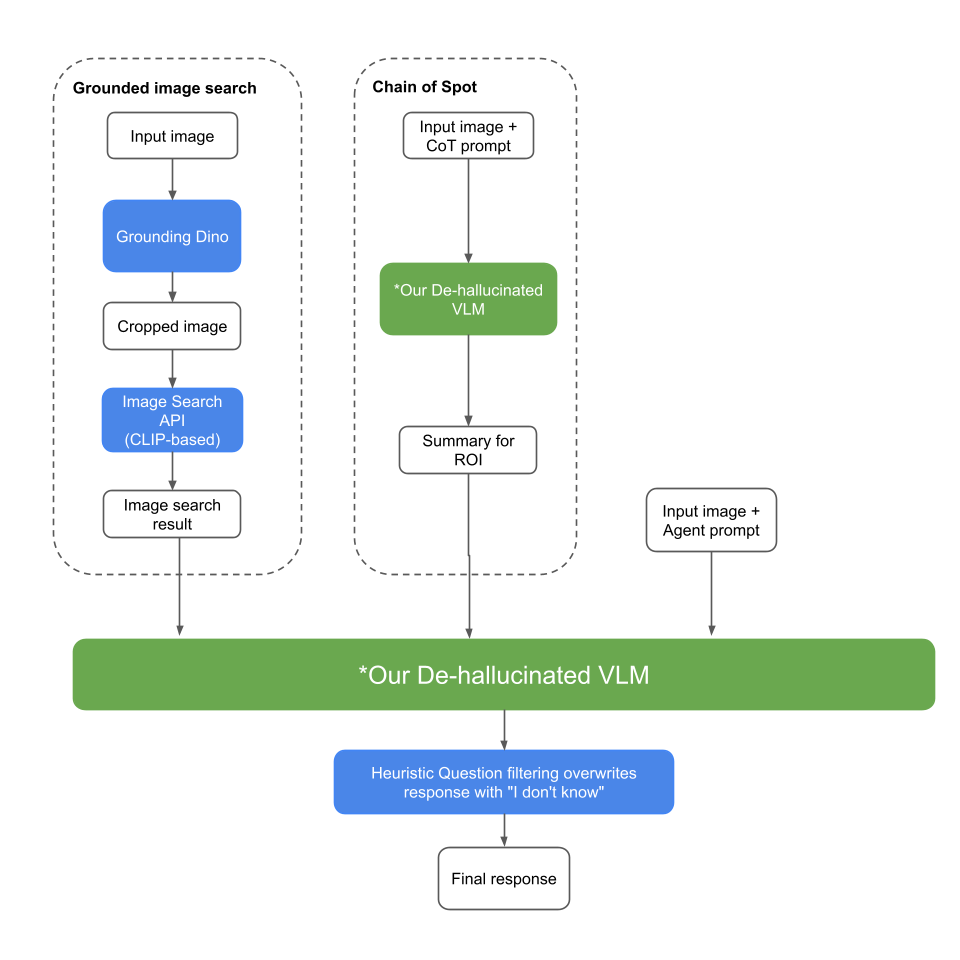}
    \caption{Overview of the System Design}
    \label{fig:overall_system_design}
  \end{subfigure}\hfill
  \begin{subfigure}{0.48\textwidth}
    \centering
    \includegraphics[width=\linewidth]{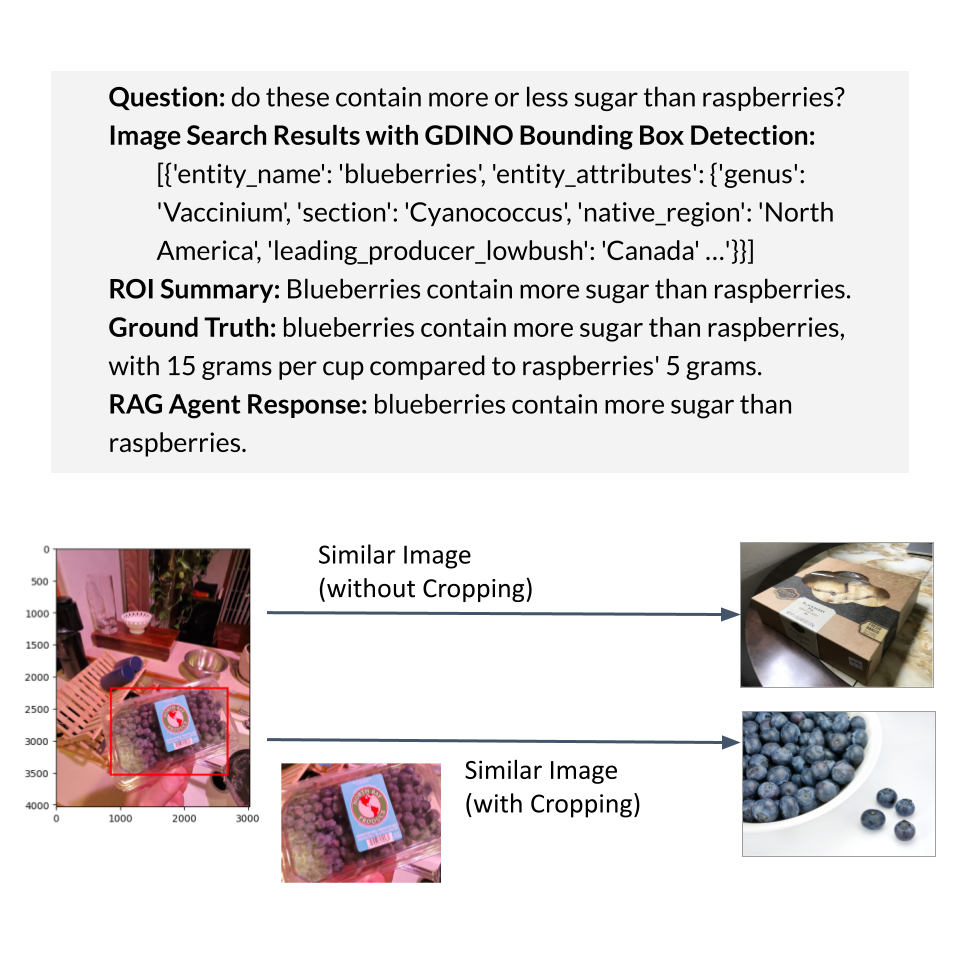}
    \caption{Example of the System Architecture in Use}
    \label{fig:architecture_example}
  \end{subfigure}
  \caption{System design and an example architecture}
  \label{fig:system_design_combined}
\end{figure*}

To address the problem discussed previously, we explored extending the capabilities of the existing VLM by introducing two key functionalities: (1) localizing the region of interest (ROI), and (2) de-hallucinating uncertain answers. These enhancements are aimed at improving the robustness and accuracy of our RAG agent (Please refer to Figure~\ref{fig:overall_system_design} for the overall system design). 

Our main system design follows a Retrieval-Augmented Generation (RAG) framework composed of the following components:
\begin{itemize}
    \item \textbf{Vision-Language Model (VLM):} We experimented with several models including BLIP, QWen, and LLaMA 3.2, ultimately selecting LLaMA 3.2 for its performance and compatibility.
    \item \textbf{Region of Interest Proposer:} A module responsible for identifying the object or region most relevant to answering the question.
    \item \textbf{Image-Based Information Retriever:} Performs web-based or local database search using cropped image regions as queries.
\end{itemize}

In addition, we fine-tune the VLM to reduce hallucination in uncertain scenarios. As discussed in the Results section, hallucination remains a significant bottleneck to overall performance.

While each module—the VLM, ROI proposer, and retriever—can be individually improved, our current focus is on enhancing the Region of Interest Proposer. Preliminary results show that models often generate irrelevant answers based on distracting background elements. Improving object localization is therefore a promising direction. Future work may explore optimization of the VLM and retriever components.

\begin{figure*}[htbp]
  \centering  \includegraphics[width=0.6\linewidth]{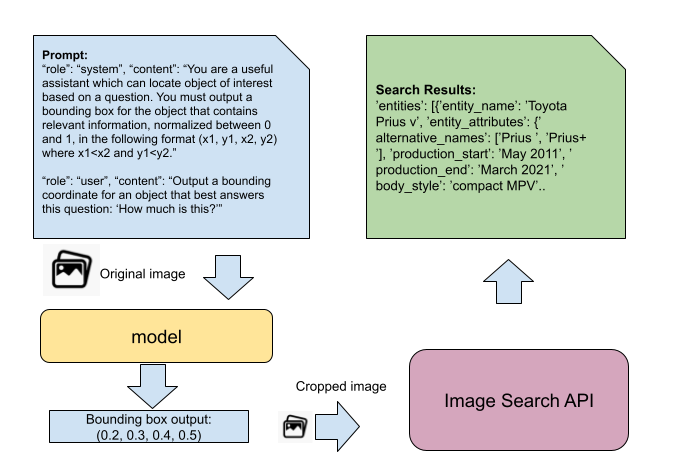}
  \caption{Overview of the Retrieval-Augmented Generation (RAG) Pipeline.}
  \label{fig:RAG}
\end{figure*}

\subsection{Localizing Region of Interest}
We reformulate visual understanding as a text-guided object localization problem \cite{li2022grounded}. Given a question, the model is tasked with identifying the most relevant object by outputting a bounding box around it. This localized ROI is then used to crop the image, reducing background noise and enabling a more focused retrieval process. The retrieved content is passed back to the model to generate the final answer (see Figure~\ref{fig:RAG}). We explore different approaches to make our RAG agent region-aware:

\subsubsection{Using a Pretrained Localizer}
Although integrating a trained localization component keeps the system self-contained, it incurs training costs and complexity. As an alternative, we also experimented with incorporating a pretrained localizer, such as Grounding DINO \cite{liu2024grounding}, to identify regions of interest prior to retrieval. The cropped region is then used to perform visual search, and the resulting information is fed back into the VLM to answer the original question.

\subsubsection{Training Vision-Language Models for Grounding via IoU Optimization}

To validate the hypothesis that a large vision-language model (LVLM) can be trained to perform spatial grounding—specifically, by accurately predicting bounding boxes from natural language prompts—we developed a four-stage fine-tuning pipeline based on the BLIP-2 architecture \cite{li2023blip}. This setup progressively unlocks more of the model's capacity for localization, culminating in a best training Intersection over Union (IoU) of \textbf{0.4552} and a final evaluation IoU of \textbf{0.4473}.

\[
\text{IoU} = \frac{|\text{Prediction} \cap \text{GroundTruth}|}{|\text{Prediction} \cup \text{GroundTruth}|}
\]

\paragraph{Training Data.}
We use the RefCOCOg dataset \cite{mao2016generation}, which consists of images paired with natural language referring expressions and annotated bounding boxes. For this experiment, we curated a subset of 5000 samples for training and 500 for final evaluation. This proof-of-concept study emphasizes clarity and reproducibility over dataset scale. We believe that expanding the dataset and unfreezing more vision layers will yield substantial performance gains.

\paragraph{Training Procedure.}
Our method consists of four training stages:

\begin{itemize}
    \item \textbf{Stage 1:} Train only the BBox prediction head, keeping BLIP-2 frozen.
    \item \textbf{Stage 2:} Unfreeze the Q-Former for co-training with the BBox head.
    \item \textbf{Stage 3:} Unfreeze the final 6 layers of the vision encoder.
    \item \textbf{Stage 4:} Use advanced optimization with 8 vision layers unfrozen and a larger dataset.
\end{itemize}

\begin{figure*}[htbp]
  \centering  \includegraphics[width=0.6\linewidth]{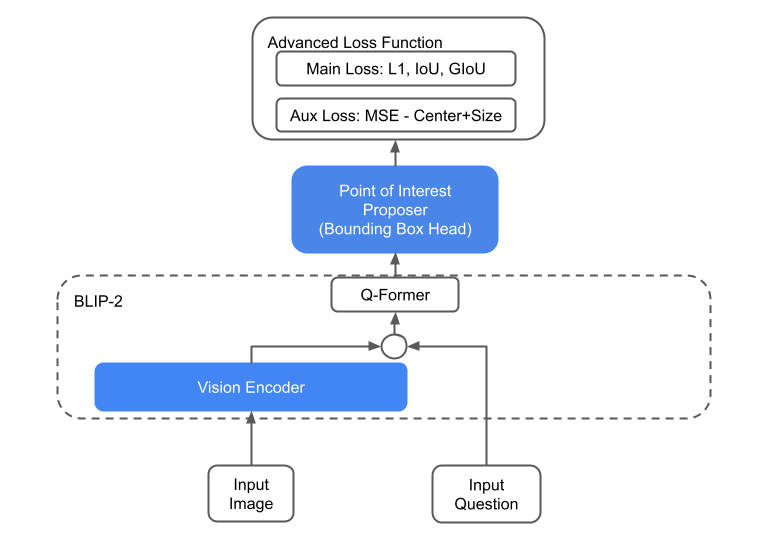}
  \caption{Architecture of the Proof-of-Concept VLM for Grounding}
  \label{fig:trained_model}
\end{figure*}

Each stage incrementally improved the model's grounding capability, confirming the feasibility of IoU-guided fine-tuning for vision-language models. (See Figure \ref{fig:trained_model}).

\paragraph{Grounding DINO vs. LVLMs.}
While models such as Grounding DINO demonstrate excellent grounding performance, they fall under the category of \textit{open-vocabulary object detection}. These models leverage pre-trained vision encoders and text embeddings to match visual regions to language descriptions, often using contrastive objectives. However, they are \textit{not large-scale vision-language models} in the generative or instruction-following sense.

In contrast, our goal is to study how to \textbf{enable grounding capabilities directly within LVLMs}, such as BLIP-2, by explicitly optimizing for spatial localization tasks (e.g., IoU prediction). This approach integrates grounding into the model's multimodal reasoning process, laying the foundation for fully unified, end-to-end instruction-following systems with spatial understanding.

\paragraph{Comparison with Other LVLMs.}
DeepSeek-VL 2~\cite{deepseekvl2} has shown competitive results on grounding benchmarks. It introduces a \lstinline|<||grounding|>
token and grounding-aware mode that help align visual and textual inputs for localization. In contrast, general-purpose LVLMs like LlaMA3.2 might not have been pretrained on bounding box data, making them less suitable for grounding tasks without sufficient training data. According to a recent survey, there has been a notable trend toward developing Grounding Multimodal Large Language Models (GMLLMs) since 2023 \cite{xiao2024towards}.

\textbf{IoU Progression Across Training Stages}

\begin{table}[h]
\centering
\begin{tabular}{|l|c|}
\hline
\textbf{Training Stage} & \textbf{Best IoU} \\
\hline
Stage 1: BBox Head Only & 0.2585 \\
Stage 2: + Q-Former & 0.3103 \\
Stage 3: + Vision Layers (6) & 0.3718 \\
Stage 4: Advanced Optimization (8 Vision Layers) & \textbf{0.4552} \\
Final Evaluation (500 samples) & \textbf{0.4473} \\
\hline
\end{tabular}
\caption{IoU scores at different stages of training for BLIP-2 on RefCOCOg.}
\end{table}

These results validate the feasibility of training LVLMs such as BLIP-2 for spatial grounding. With additional scale in data, model size, and fine-tuning depth, we expect future iterations to cross the 0.5 IoU threshold required for high-quality object localization. Due to compute limitation and the fact that Llama 3.2 was not explicitly pre-trained on grounding tasks for bounding box generation, we decided to use Grounding DINO for object localization, while finetuning models for localization is left for future work.

\subsection{Finetuning to Reduce Hallucination}
We also fine tuned the model to say "I don't know" in some cases to reduce hallucination. For example, for the following question in Table \ref{tab:eval-00a48}, Llama-3.2-Vison-11B outputs an incorrect answer. We can train the model to answer "I don't know" for "who" type question, since this typically requires external knowledge. We investigate the questions based on the type shown in Figure \ref{fig:accuracy_by_question_type}, and then fine tune the model to only answer the questions with high confidence.

\newcolumntype{L}{>{\raggedright\arraybackslash}X}   

\begin{table*}[t]
  \caption{Evaluation record for Interaction ID \texttt{00a48}}
  \label{tab:eval-00a48}
  \centering
  \begin{tabularx}{\textwidth}{@{}
    p{0.08\textwidth}  
    p{0.15\textwidth}  
    p{0.40\textwidth}  
    p{0.25\textwidth}  
    p{0.10\textwidth}  
    @{}}
    \toprule
    \textbf{ID} & \textbf{Query} & \textbf{Agent Response} & \textbf{Ground Truth} & \textbf{Result} \\
    \midrule
    00a48 &
    Who invented this kind of tape? &
    The tape measure was invented by the French tailor \emph{Pierre-Frédéric Guillaume}, who patented the first practical retractable tape measure in 1829 (prototype in 1821). &
    James Chesterm... &
    \textsc{Incorrect} \\
    \bottomrule
  \end{tabularx}
\end{table*}

\section{Results}
\label{sec:results}

\subsection{Dataset and Task Definition}

We analyzed a VQA dataset collected from Meta RayBan smart glasses \cite{meta_crag_mm_challenge_2025}, which contains both single-turn and multi-turn image-question-answer pairs across 14 diverse domains, including shopping, food, and science. We primarily evaluate on the v2 version of this dataset, which contains 1938 single-turn questions. For comparison, we also report results on the earlier v1 version, which includes 1548 questions, during our ablation studies.

\begin{figure}[h]
    \centering
    \begin{subfigure}[b]{0.2\textwidth}
        \centering
        \includegraphics[width=\linewidth]{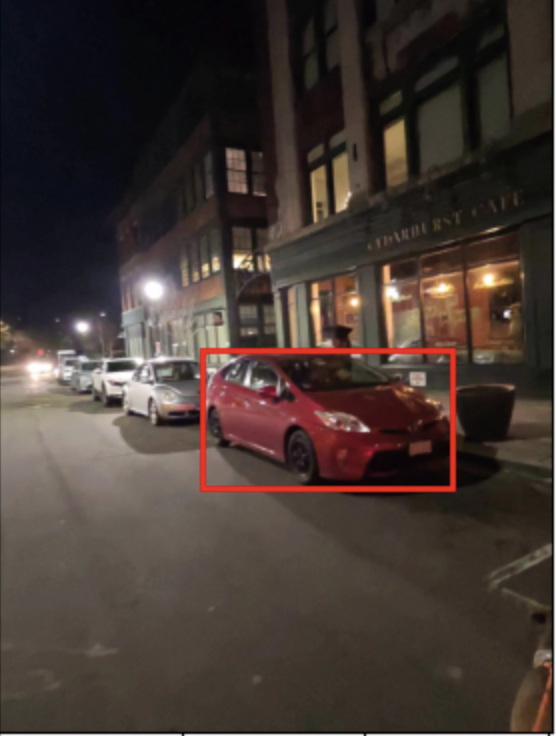 }
        \caption{Input image and region of interest}
        \label{fig:image1}
    \end{subfigure}
    \hspace{0.05\textwidth}
    \begin{subfigure}[b]{0.2\textwidth}
        \centering
        \includegraphics[width=\linewidth]{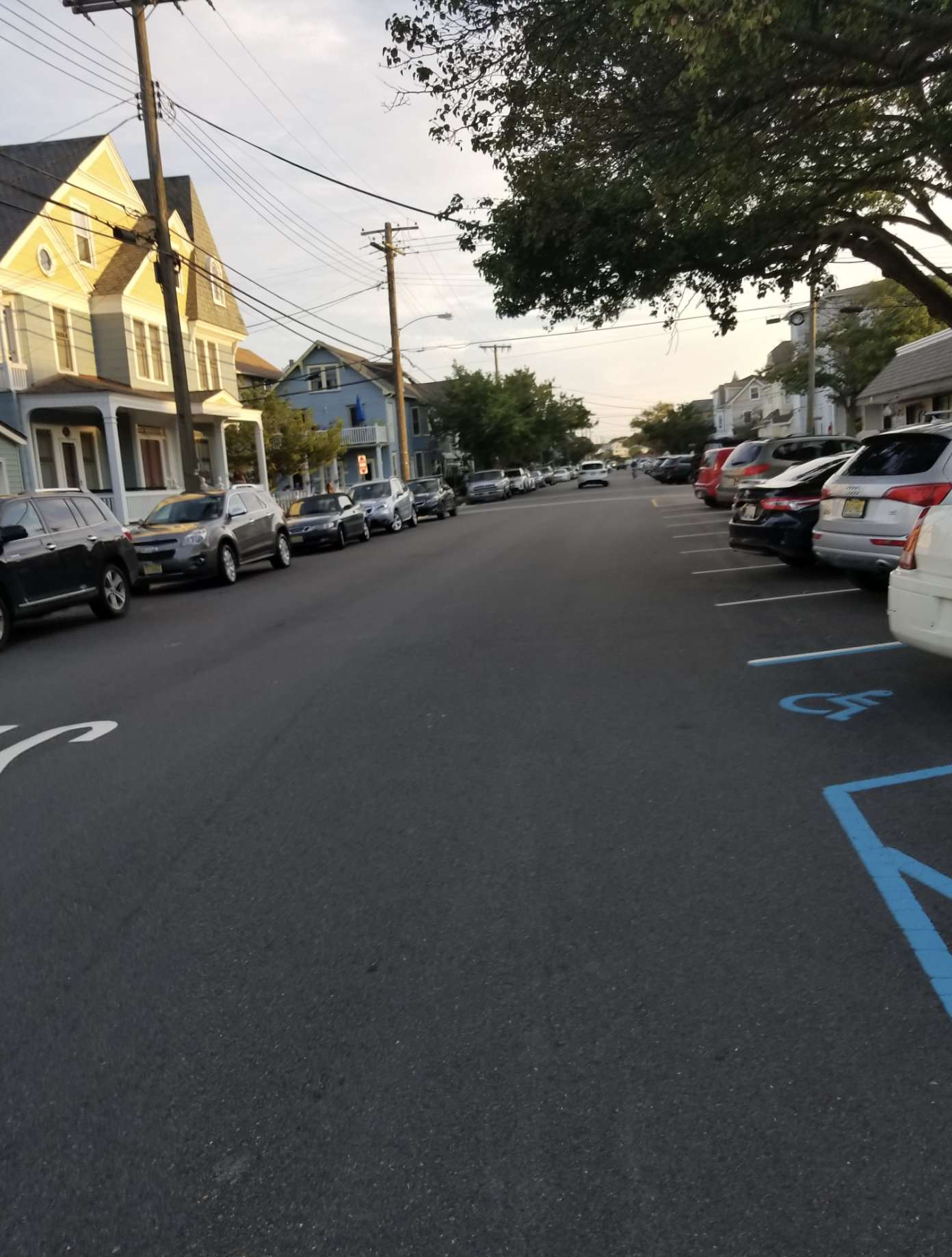}
        \caption{Retrieved image doesn't match the object in question}
        \label{fig:image2}
    \end{subfigure}
    \caption{Example image retrieval without text grounding}
    \label{fig:side-by-side}
\end{figure}

Used for evaluation of our work, this dataset presents a significant challenge due to its varying image quality and ambiguous questions, requiring models to extract the most relevant information from noisy inputs. For instance, consider an image showing several cars near buildings. Performing image retrieval directly on the full image would likely yield results focused on street scenes or buildings, since large background elements tend to dominate the image. Therefore, if the question is “How many passengers can the red car seat?”, a retrieval system unaware of the object of interest will fail to provide accurate information (see Figure \ref{fig:side-by-side}).

\begin{figure}[htbp]
  \centering
  \begin{subfigure}{0.9\linewidth}
    \centering
    \includegraphics[width=\linewidth]{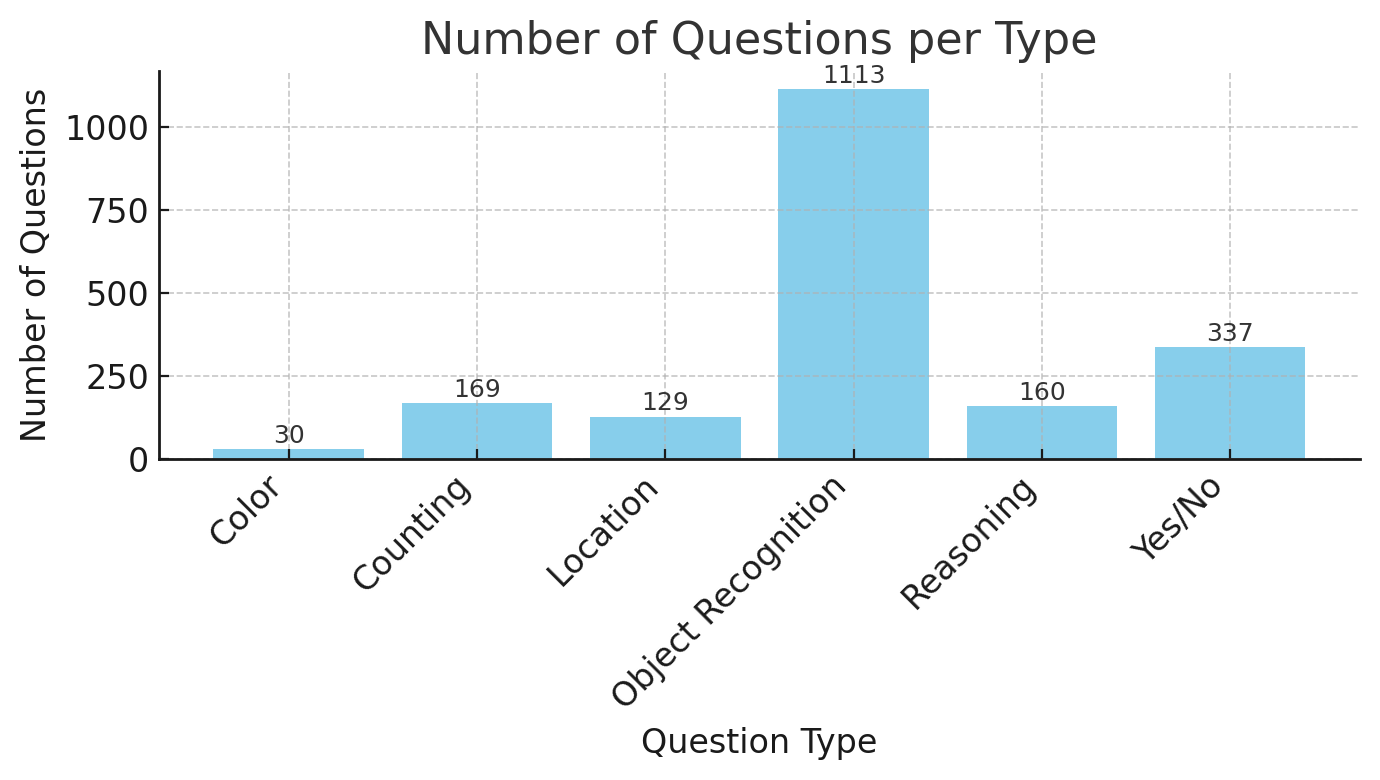}
    \caption{Question‐type count}
    \label{fig:question_type_count}
  \end{subfigure}

  \vspace{1em}

  \begin{subfigure}{0.9\linewidth}
    \centering
    \includegraphics[width=\linewidth]{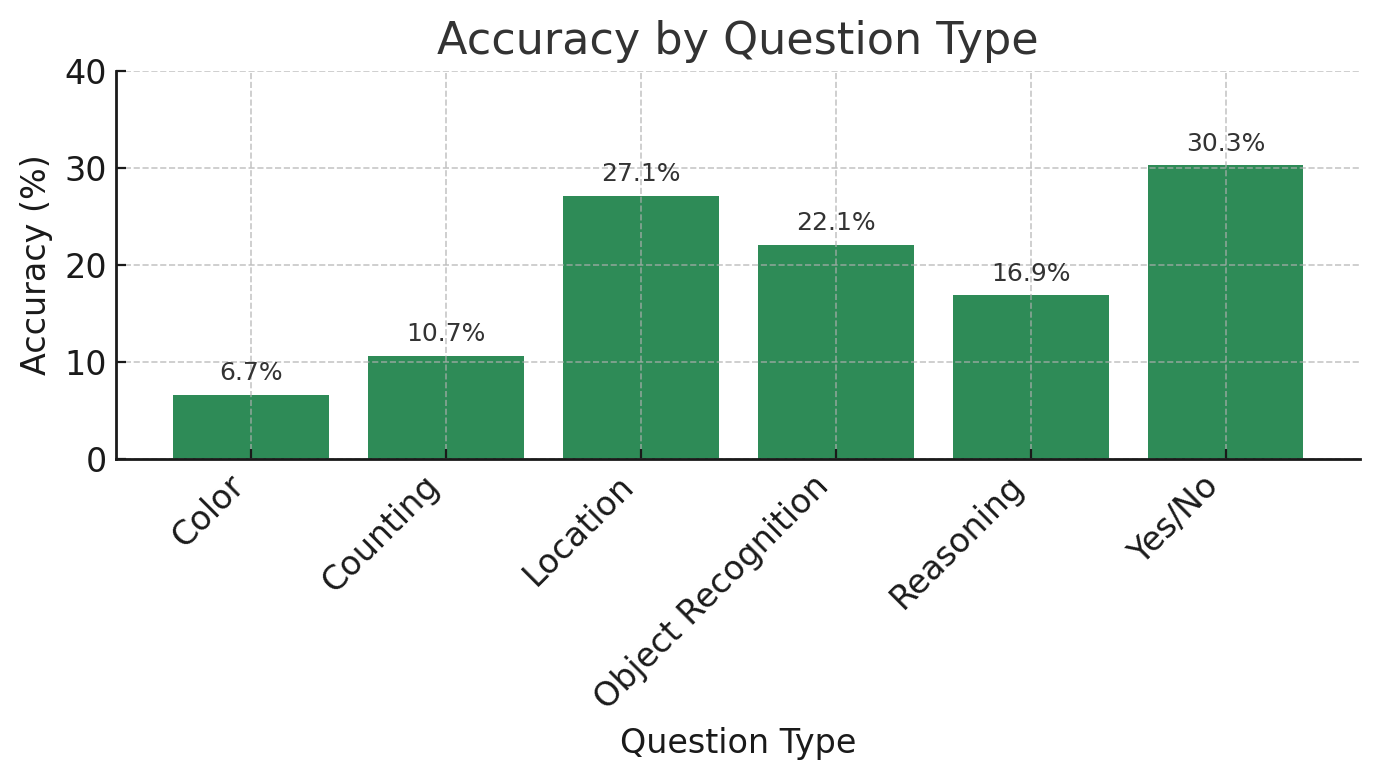}
    \caption{Accuracy by question type}
    \label{fig:accuracy_by_question_type}
  \end{subfigure}

  \caption{Analysis by question type}
  \label{fig:question_type_analysis}
\end{figure}

The dataset features a variety of question types, such as color, counting, location, object recognition, reasoning, and yes/no queries. According to the distribution of question types in the v2 dataset (see Figure \ref{fig:question_type_count}), object recognition questions constitute the largest portion. This skew towards object recognition highlights the need for models to possess strong, text-grounded visual understanding, and sometimes be able to localize the object of interest among multiple other objects in the background based on the input text.

This dataset presents a realistic and challenging benchmark for vision-language systems, making it a strong candidate for evaluating model performance. In the following subsection, we present baseline results using several state-of-the-art VLMs.

\subsubsection{Image Search API}

We utilize a prebuilt image search API provided by Meta \cite{jyotish_search_api_2025}, which contains a database of 900K images and associated metadata. This API converts images into embeddings using the CLIP model \cite{radford2021learning}, retrieves similar images based on cosine similarity, and returns them with relevant information. We observed that specifying the region of interest (ROI) is crucial for retrieval-augmented generation (RAG), since the retrieved results differ significantly depending on the ROI. When unrelated regions are used, the retrieved information does not aid the model in answering the question and may even promote hallucinations. For example, as shown in Listing \ref{lst:no_roi}, using the full image from the previous section results in completely irrelevant retrieval when the user’s actual question concerns the car parked on the street rather than the street itself.

\FloatBarrier

\begin{lstlisting}[caption={Example Cropped Region Search Result},label={lst:cropped}]
'entities': [{
  'entity name': 'Toyota Prius v',
  'entity attributes': {
    'alternative names': ['Prius alpha','Prius+'],
    'production start': 'May 2011',
    'production end': 'March 2021',
    'body style': 'compact MPV'
  }
}]
\end{lstlisting}

\begin{lstlisting}[caption={Example Image Search Without Knowing Region Of Interest},label={lst:no_roi}]
'entities': [{
  'entity_name': 'Ocean Grove, New Jersey',
  'entity_attributes': {
    'official_name': 'Ocean Grove, New Jersey',
    'settlement_type': '[Census-designated place]',
    'image_skyline': 'Ocean_Grove_Welcome_Sign.jpg',
    'imagesize': '250x200px',
    'image_caption': 'Ocean Grove welcome sign'
  }
}]
\end{lstlisting}

\subsection{Baseline Model Performance}
We have collected baseline performance for three popular open source VLMs: BLIP (Bootstrapping Language-Image Pre-training) \cite{li2022blip} , Llama 3.2 (11B) \cite{grattafiori2024llama3herdmodels} and Qwen 2.5 (3B) \cite{bai2025qwen25vltechnicalreport}, on this dataset without any information retrieval implemented.

\subsection{End To End Evaluation Method}
We use \texttt{GPT-4o-mini} as a judge, guided by three rules: (i) a prediction is correct when it contains all key information in the ground truth, (ii) paraphrasing is acceptable if the meaning is unchanged, and (iii) a prediction is incorrect if it introduces errors or omits essentials. For each question we assign a scalar score—\textbf{Perfect} (1.0), \textbf{Acceptable} (0.5, minor non-harmful flaws), \textbf{Missing} (0.0, refusal/``I~don’t know’’), or \textbf{Incorrect} (–1.0, wrong or irrelevant).  A system’s \emph{Truthfulness Score} is the mean of these values across the evaluation set, yielding a range of –1 (all wrong) to 1 (all perfect). Listing \ref{lst:judge} shows the full prompt for the judge.

\subsection{Results on Vanilla Vision-Language Models} Based on the preliminary results of three unmodified models pre-trained on 1,548 questions from \autoref{tab:vlm-overall}, the blip-vqa-base model performs poorly in the real-world data set with only accuracy of 3. 49\%. The Qwen-VL-2.5-3B and Llama-3.2-Vision-11B perform relatively better with 18.09\% and 26.23\% accuracy. The Hallucination rate is relatively higher in Llama-3.2-Vision-11B compared to Qwen-VL-2.5-3B. As a result, Llama-3.2-Vision-11B has the highest truthfulness score, and Qwen-VL-2.5-3B ranks the second. This will be used as a baseline in future experiments.

\subsection{Ablation Study}
\subsubsection{Baseline on V1 Dataset}
The CRAG MM dataset has two versions, where the first version v1 includes 1548 single turn questions, and the second version v2 includes 1938 single turn questions. We evaluated our baseline on v1 dataset with three different pre-trained models: BLIP, Qwen-VL-2.5-3B and Llama-3.2-Vision-11B. Since Llama-3.2-Vision-11B performs relatively great, we use Llama-3.2-Vision-11B for our customized solutions and evaluate only Llama-3.2-Vision-11B on v2 dataset as baseline for our customized solutions.

Based on the results of three pretrained unmodified models on v1 1548 questions from \autoref{tab:vlm-overall}, the blip-vqa-base model performs poorly on the real world data set with only 3.49\% accuracy. The Qwen-VL-2.5-3B and Llama-3.2-Vision-11B perform relatively better with 18.09\% and 26.23\% accuracy. The Hallucination rate is relatively higher in Llama-3.2-Vision-11B compared to Qwen-VL-2.5-3B. As a result, Llama-3.2-Vision-11B has the highest truthfulness score, and Qwen-VL-2.5-3B ranks the second.

\subsubsection{Localization}
We compare two approaches to incorporating region-of-interest localization into our retrieval-augmented generation (RAG) system. The first uses Grounding DINO, a pretrained object localizer, to detect and crop the most relevant image region for answering the question. The second follows a Chain-of-Spot-style prompting strategy, where the model is first asked to describe the region of interest before retrieving and answering. While both approaches aim to improve retrieval quality by focusing on relevant image content, they differ in whether localization is model-guided (via prompting) or external (via a pretrained detector).

We evaluated our RAG agent on the new version of the dataset using two different prompting strategies. Both experiments use Grounding DINO (grounding-dino-tiny, 172M) for bounding box extraction instead of our internally trained bounding box heads due to limited compute budget.

Table~\ref{tab:rag-results} compares the performance of the baseline model (without RAG) against the two prompting approaches. In the first strategy, we crop the input image using the detected bounding box, perform a retrieval based on the cropped region, and use the retrieved information to answer the question. In the second strategy, we follow a Chain-of-Spot-style (CoS) approach \cite{dong2024chainofspot}: the model first summarizes the region of interest, then we crop the image using Grounding DINO, and finally feed both the summary and search results back into the model to answer the original question.

Surprisingly, using Grounding DINO alone in the first experiment led to a nearly 5\% drop in accuracy, with a hallucination rate comparable to the baseline. In contrast, the Chain-of-Spot-style prompting improved accuracy beyond the baseline, but introduced a higher hallucination rate (5\%), which ultimately lowered the overall score.

Upon analyzing outputs from different stages of our RAG agent, we observed the following:

\begin{itemize}
  \item When image-based information retrieval returns completely irrelevant content, it can mislead the model into producing incorrect answers (Table~\ref{tab:not-useful-search}). This limits the effectiveness of RAG and results in performance comparable to the baseline. In contrast, Chain-of-Spot prompting mitigates this issue by first asking the model to describe what it sees before incorporating retrieved information. We hypothesize that this approach encourages the model to rely more confidently on its own visual understanding than on potentially misleading external sources, as we have seen the model output the exact same answer as the summary, ignoring RAG results. 
  
  \item However, Chain-of-Spot prompting can also increase hallucination. That is, once the model identifies the region of interest, it may become overly confident in its predictions. As shown in Table~\ref{tab:chain-of-spot-over-confident}, this can sometimes override correct prior knowledge, leading to confidently incorrect answers. This behavior contributes to a higher hallucination rate and lowers the overall accuracy. This is evident given the low missing rate and the model is less likely to output "I don't know".

  \item Computation wise, the gounding DINO is extremely efficient to generate bounding boxes, while the Chain-of-Spot prompting requires the model to look at the image twice, leading to almost double inference time. On A100 with 80 GB VRAM using a batch size of 36, one full evaluation on v2 dataset takes 1 hour, and on average each batch takes 1.8 seconds. This is consistent with other test-time scaling technique like Chain-of-Thought \cite{wei2022chainofthought}.
\end{itemize}

\subsubsection{De-hallucination with Fine-tuning Model}
While accuracy improved with more advanced prompting strategies, hallucination rates also increased. To address this, we fine-tuned the model to condition its responses on the question type. This enabled it to abstain from answering difficult reasoning questions when uncertain, effectively generating "I don't know" responses instead of hallucinations.

As shown in Table~\ref{tab:llama32v11b-comparison-de-hallucination}, without using RAG, this approach reduced the hallucination rate from 65.79\% to 19.14\%, and improved the truthfulness score from $-0.4360$ to $-0.0738$. This significant improvement demonstrates the impact of hallucination control on downstream performance.

Since our primary focus is to study the effectiveness of the prediction with grounding information, we did not overengineer tricks on reducing the hallucination rate to purely improving the truthfulness score. That's been said, there is clear path to decrease hallucination further with this methodology.

\subsubsection{De-hallucination with Image Search Threshold}
We also applied an additional image search threshold in the final GroundSight solution. The final solution has a threshold of 0.75. Setting a CLIP similarity threshold of 0.75 filters irrelevant data while preserving semantically grounded matches.

\subsubsection{End to end performance}
After analyzing localization and de-hallucination strategies independently, we integrated them into our full GoundSight agent and compare against individual techniques in Table ~\ref{tab:groundsight-results}. We observed that they act complementarily: localization enhances answer relevance by grounding the model’s attention, Chain-of-Spot encourages the model to be more confident in it's own knowledge to avoid being misled by irrelevant RAG results, while de-hallucination ensures cautiousness when the model is uncertain. This balance improves overall performance by reducing confidently wrong answers and encouraging selective abstention on harder questions. Our agent achieved the highest truthfulness score of -0.049.

\vspace{1em}
\noindent
In summary, grounding and de-hallucination each contribute to improved VQA performance, and their combination in GroundSight offers a practical path toward more accurate and trustworthy answers.

\section{Conclusion}
In summary, this work demonstrates that incorporating text-grounded object localization into retrieval-augmented VQA systems enables models to produce more accurate and context-aware answers by focusing on the most relevant image regions for each question. By leveraging Chain-of-Spot-style prompting, our RAG agent is able to effectively combine retrieved web content with the model’s own visual understanding. Experiments on challenging real-world datasets show that this localization-based strategy improves accuracy over baseline methods, though at the cost of increased hallucination.

Interestingly, our de-hallucinated model, which more frequently responds with ``I don't know,'' achieves the highest overall score—\allowbreak highlighting a valuable real-world insight: providing an incorrect answer can be more detrimental than admitting uncertainty. This underscores a practical challenge of using RAG agents in safety-critical applications. 

Future work could explore combining Chain-of-Spot prompting with a fine-tuned, de-hallucinated VLM, enabling the agent to retain low hallucination rates while still leveraging external information to enhance accuracy.

\appendix
\section{Additional Results}

\begin{figure*}[t]
\centering
\begin{minipage}{\textwidth}
\begin{lstlisting}[
  caption={Agent prompt},
  label={lst:judge},
  language=Python,
  breaklines=true,
  frame=single,
  numbers=left,
  basicstyle=\ttfamily\small,
  captionpos=b
]
SYSTEM PROMPT: "You are a precise and cautious assistant that truthfully answers user questions about the provided image augmented with online search information. Only answer if you are confident and have the necessary knowledge. If you are not absolutely certain about the answer, reply with exactly: 'I don't know', without any further explanation. Do not use any other phrases like 'I don't have details', 'It depends', or 'I don't have enough information'. Your response must be concise and must not exceed 75 words."}
USER PROMPT: "Context that may be relevant to the objects in question:\n"
+ search_results + "Answer this question: " + query}

\end{lstlisting}
\end{minipage}
\end{figure*}

\begin{figure*}[t]
\centering
\begin{minipage}{\textwidth}
\begin{lstlisting}[
  caption={Chain Of Spot prompt},
  label={lst:judge},
  language=Python,
  breaklines=true,
  frame=single,
  numbers=left,
  basicstyle=\ttfamily\small,
  captionpos=b
]
SYSTEM PROMPT: "You are a helpful assistant that can summarize a region of interest of the image based on user's question. The summary should be concise and only contain a simple description that must not exceed 10 words. The summary must not answer the question."
USER PROMPT: "Provide a concise summary for object of interest that can answer the following question: '" + query + "'"}
\end{lstlisting}
\end{minipage}
\end{figure*}

\begin{figure*}[t]
\centering
\begin{minipage}{\textwidth}
\begin{lstlisting}[
  caption={LLM judge prompt},
  label={lst:judge},
  language=Python,
  breaklines=true,
  frame=single,
  numbers=left,
  basicstyle=\ttfamily\small,
  captionpos=b
]
"You are an expert evaluator for question answering systems. "
"Your task is to determine if a prediction correctly answers a question based on the ground truth.\n\n"
"Rules:\n"
"1. The prediction is correct if it captures all the key information from the ground truth.\n"
"2. The prediction is correct even if phrased differently as long as the meaning is the same.\n"
"3. The prediction is incorrect if it contains incorrect information or is missing essential details.\n"
"Output a JSON object with a single field 'accuracy' whose value is true or false."
\end{lstlisting}
\end{minipage}
\end{figure*}

\begin{table*}[htbp]
  \centering
  \caption{Overall evaluation metrics for Llama-3.2-Vision-11B on V1 Dataset (Baseline vs.\ De-hallucination)}
  \label{tab:llama32v11b-comparison-de-hallucination}
  \setlength{\tabcolsep}{5pt}
  \renewcommand{\arraystretch}{1.2}
  \begin{tabular}{l p{1.5cm} p{1.5cm} p{1.5cm} p{1.5cm} p{1.6cm} p{1.8cm} p{2cm}}
    \toprule
    \textbf{Model} &
    \textbf{Total conv.} &
    \textbf{Total turns} &
    \textbf{Exact acc. (\%)} &
    \textbf{Accuracy (\%)} &
    \textbf{Missing rate (\%)} &
    \textbf{Hallucination rate (\%)} &
    \textbf{Truthfulness score} \\
    \midrule
    Llama-3.2-Vision-11B (Baseline)         & 1548 & 1548 & 0.84 & 26.23 & 13.24 & 60.53 & $-0.3430$ \\
    Llama-3.2-Vision-11B (De-hallucination) & 1548 & 1548 & 0.90 & 12.60 & 69.64 & 17.76 & $-0.0517$ \\
    \bottomrule
  \end{tabular}
\end{table*}

\begin{table*}[htbp]
  \centering
  \caption{Overall evaluation metrics for three vision--language models without RAG}
  \label{tab:vlm-overall}
  \setlength{\tabcolsep}{5pt}
  \renewcommand{\arraystretch}{1.2}
  \begin{tabular}{l p{1.5cm} p{1.5cm} p{1.5cm} p{1.5cm} p{1.6cm} p{1.8cm} p{2cm}}
    \toprule
    \textbf{Model} & \textbf{Total conv.} & \textbf{Total turns} & \textbf{Exact acc. (\%)} & \textbf{Accuracy (\%)} & \textbf{Missing rate (\%)} & \textbf{Hallucination rate (\%)} & \textbf{Truthfulness score} \\
    \midrule
    blip-vqa-base & 1548 & 1548 & 0.00 & 3.49 & 0.00 & 96.51 & $-0.9302$ \\
    Qwen-VL-2.5-3B & 1548 & 1548 & 0.78 & 18.09 & 33.72 & 48.19 & $-0.3010$ \\
    Llama-3.2-Vision-11B & 1548 & 1548 & 0.84 & 26.23 & 13.24 & 60.53 & $-0.3430$ \\
    \bottomrule
  \end{tabular}
\end{table*}

\begin{table*}[htbp]
  \centering
  \caption{Overall evaluation metrics for Llama-3.2-Vision-11B on V2 Dataset (Baseline vs.\ De-hallucination)}
  \label{tab:llama32v11b-comparison-de-hallucination}
  \setlength{\tabcolsep}{5pt}
  \renewcommand{\arraystretch}{1.2}
  \begin{tabular}{l p{1.5cm} p{1.5cm} p{1.5cm} p{1.5cm} p{1.6cm} p{1.8cm} p{2cm}}
    \toprule
    \textbf{Model} & \textbf{Total conv.} & \textbf{Total turns} & \textbf{Exact acc. (\%)} & \textbf{Accuracy (\%)} & \textbf{Missing rate (\%)} & \textbf{Hallucination rate (\%)} & \textbf{Truthfulness score} \\
    \midrule
    Llama-3.2-Vision-11B (Baseline) & 1938 & 1938 & 0.46 & 22.19 & 12.02 & 65.79 & $-0.4360$ \\
    Llama-3.2-Vision-11B (De-hallucination) & 1938 & 1938 & 0.67 & 11.76 & 69.09 & 19.14 & $-0.0738$ \\
    \bottomrule
  \end{tabular}
\end{table*}

\begin{table*}[htbp]
  \centering
  \caption{Overall evaluation metrics for RAG implementation}
  \label{tab:rag-results}
  \setlength{\tabcolsep}{5pt}
  \renewcommand{\arraystretch}{1.2}
  \begin{tabular}{l p{1.5cm} p{1.5cm} p{1.5cm} p{1.5cm} p{1.6cm} p{1.8cm} p{2cm}}
    \toprule
    \textbf{Model} & \textbf{Total conv.} & \textbf{Total turns} & \textbf{Exact acc. (\%)} & \textbf{Accuracy (\%)} & \textbf{Missing rate (\%)} & \textbf{Hallucination rate (\%)} & \textbf{Truthfulness score} \\
    \midrule
    Llama-3.2-Vision-11B (Baseline no RAG) & 1938 & 1938 & 0.46 & 22.19 & 12.02 & 65.79 & $-0.4360$ \\
    GDINO & 1938 & 1938 & 0.15 & 17.75 & 16.87 & 65.38 & $-0.4763$ \\
    GDINO + chain of spot & 1938 & 1938 & 0.83 & 25.64 & 3.77 & 70.59 & $-0.4494$ \\
    \bottomrule
  \end{tabular}
\end{table*}

\begin{table*}[htbp]
  \centering
  \caption{Overall evaluation metrics for RAG implementation}
  \label{tab:groundsight-results}
  \setlength{\tabcolsep}{5pt}
  \renewcommand{\arraystretch}{1.2}
  \begin{tabular}{l p{1.5cm} p{1.5cm} p{1.5cm} p{1.5cm} p{1.6cm} p{1.8cm} p{2cm}}
    \toprule
    \textbf{Model} & \textbf{Total conv.} & \textbf{Total turns} & \textbf{Exact acc. (\%)} & \textbf{Accuracy (\%)} & \textbf{Missing rate (\%)} & \textbf{Hallucination rate (\%)} & \textbf{Truthfulness score} \\
    \midrule
    \rowcolor{gray!15}
    Llama-3.2-Vision-11B (Baseline no RAG) & 1938 & 1938 & 0.46 & \textbf{22.19} & 12.02 &  \textbf{65.79} & $-0.4360$ \\
    
    \shortstack[l]{GroundSight \\+GDINO} & 1938 & 1938 & 0.15 & 17.75 & 16.87 & 65.38 & $-0.4763$ \\
    \rowcolor{gray!15}
    \shortstack[l]{GroundSight \\+GDINO + CoS} & 1938 & 1938 & 0.83 & \textbf{25.64} & 3.77 & 70.59 & $-0.4494$ \\
    \shortstack[l]{GroundSight \\
    Llama-3.2-Vision-11B (De-hallucination)} & 1938 & 1938 & 0.67 & 11.76 & 69.09 & 19.14 & $-0.0738$ \\
    \rowcolor{gray!15}
    \shortstack[l]{GroundSight \\ (De-hallucination + GDINO + CoS \\ w/ Image Search threshold = 0.75)}
    & 1938 & 1938 & 0.36 & 8.98 & 77.14 & \textbf{13.88} & $-0.0490$ \\
    \bottomrule
  \end{tabular}
\end{table*}

\begin{table*}[t]
\centering
\caption{Model output when search result is not useful}
\label{tab:not-useful-search}
\renewcommand{\arraystretch}{1.2}
\small
\begin{adjustbox}{width=\textwidth}
\begin{tabular}{|p{3cm}|p{3cm}|p{4cm}|p{4cm}|p{4cm}|}
  \hline
  \textbf{Question} & \textbf{Image} & \textbf{Image Crop + Search} & \textbf{Chain-of-Spot} & \textbf{Baseline} \\
  \hline
  What is the typical filling of this Chinese steamed bun? &
  \begin{minipage}[t]{\linewidth}
    \textbf{Original:}\\
    \includegraphics[width=\linewidth]{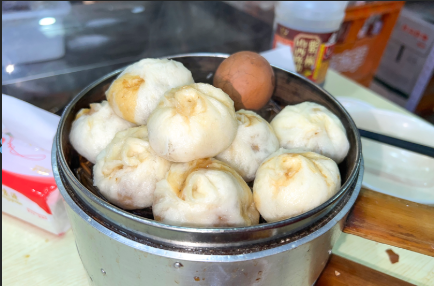}\\[0.5ex]
    \textbf{Cropped:}\\
    \includegraphics[width=\linewidth]{img/steam-bun.png}
  \end{minipage} &
  \textbf{Search result:} Blood soup. \newline
  \textbf{Response (incorrect):} The typical filling of this Chinese steamed bun is not blood soup, as the image shows a steamed bun with a brown filling, not a soup. &
  \textbf{Search result:} Blood soup. \newline
  \textbf{ROI summary:} The typical filling of this Chinese steamed bun is pork. \newline
  \textbf{Response (correct):} The typical filling of this Chinese steamed bun is pork. &
  \textbf{Response (correct):} The typical filling of this Chinese steamed bun is pork. \\
  \hline
  \multicolumn{1}{|p{3cm}|}{\textbf{Groundtruth:} The typical filling is pork.} & \multicolumn{4}{l|}{} \\
  \hline
\end{tabular}
\end{adjustbox}
\end{table*}

\begin{table*}[t]
\centering
\caption{Model output when Chain-of-Spot makes overconfident predictions}
\label{tab:chain-of-spot-over-confident}
\renewcommand{\arraystretch}{1.2}
\small
\begin{adjustbox}{width=\textwidth}
\begin{tabular}{|p{3cm}|p{3cm}|p{4cm}|p{4cm}|p{4cm}|}
  \hline
  \textbf{Question} & \textbf{Image} & \textbf{Image Crop + Search} & \textbf{Chain-of-Spot} & \textbf{Baseline} \\
  \hline
  How old was this artist when he started hosting his own show on NBC? &
  \begin{minipage}[t]{\linewidth}
    \textbf{Original:}\\
    \includegraphics[width=\linewidth]{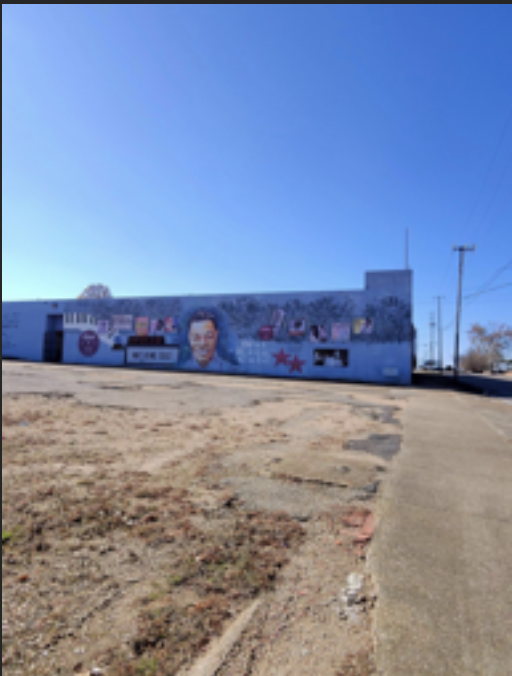}\\[0.5ex]
    \textbf{Cropped:}\\
    \includegraphics[width=\linewidth]{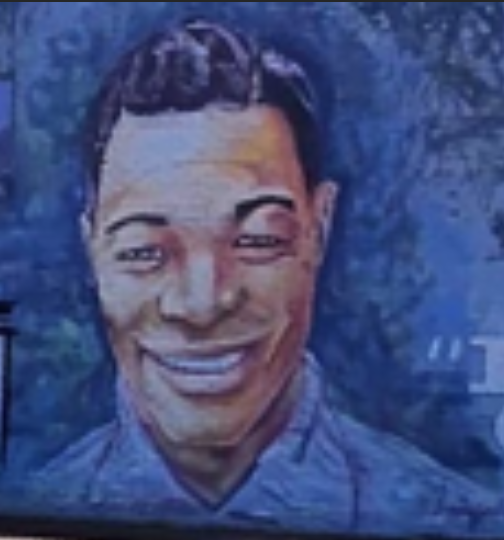}
  \end{minipage} &
  \textbf{Search result:} Levi Strauss \& Co. is an American clothing company. \newline
  \textbf{Final answer:} I don't know. &
  \textbf{Search result:} Levi Strauss \& Co. is an American clothing company. \newline
  \textbf{ROI summary:} The object of interest is a mural of Nat King Cole, an American singer and musician. \newline
  \textbf{Response (incorrect):} Nat King Cole was 31 years old when he started hosting his own show on NBC, "The Nat King Cole Show," in 1956. &
  \textbf{Response (correct):} Nat King Cole was born on March 17, 1919, and he started hosting his own show on NBC in 1956. Therefore, he was 37 years old when he started hosting his own show on NBC. \\
  \hline
  \multicolumn{1}{|p{3cm}|}{\textbf{Groundtruth:} Nat King Cole was 37 years old.} & \multicolumn{4}{l|}{} \\
  \hline
\end{tabular}
\end{adjustbox}
\end{table*}

\clearpage

\begin{acks}
We sincerely thank Sabri Eyuboglu for the valuable insights on the topic of vision-language models. We thank Qi Tang for sharing knowledge about infrastructure. We are grateful to the Department of Computer Science at Stanford University for support, including computing resources, and to the teaching staff—including Fei-Fei Li, Ehsan Adeli, Ranjay Krishna and others—for clarifying several technical points; this work builds on a prior project conducted there.
\end{acks}

\bibliographystyle{ACM-Reference-Format}
\bibliography{sample-base}


\end{document}